\documentclass[10pt,twocolumn,letterpaper]{article}

\usepackage{cvpr}
\usepackage{times}
\usepackage{epsfig}
\usepackage{graphicx}
\usepackage{amsmath}
\usepackage{amssymb}
\usepackage{float}
\usepackage{subcaption}
\usepackage{caption}

\usepackage[pagebackref=true,breaklinks=true,letterpaper=true,colorlinks,bookmarks=false]{hyperref}

\cvprfinalcopy 

\def\cvprPaperID{1877} 

\ifcvprfinal\fi
\begin{document}

\title{DA-GAN: Instance-level Image Translation by Deep Attention Generative Adversarial Networks}

\author{Shuang Ma\\
The State Universtiy of New York at Buffalo\\
{\tt\small shuangma@buffalo.edu}
\and
Jianlong Fu\\
Microsoft Research\\
{\tt\small jianf@microsoft.com}
\and 
Chang Wen Chen\\
The State Universtiy of New York at Buffalo\\
{\tt\small chencw@buffalo.edu}
\and Tao Mei\\
Microsoft Research\\
{\tt\small tmei@microsoft.com}
}

\maketitle


\begin{abstract}
Unsupervised image translation, which aims in translating two independent sets of images, is challenging in discovering the correct correspondences without  paired data.
Existing works build upon Generative Adversarial Network (GAN) such that the distribution of the translated images are indistinguishable from the distribution of the target set. However, such set-level constraints cannot learn the instance-level correspondences (e.g. aligned semantic parts in object configuration task). This limitation often results in false positives (e.g. geometric or semantic artifacts), and further leads to mode collapse problem.
To address the above issues, we propose a novel framework for instance-level image translation by Deep Attention GAN (DA-GAN). Such a design enables DA-GAN to decompose the task of translating samples from two sets into translating instances in a highly-structured latent space. 
Specifically, we jointly learn a deep attention encoder, and the instance-level correspondences could be consequently discovered through attending on the learned instance pairs. Therefore, the constraints could be exploited on both set-level and instance-level.
Comparisons against several state-of-the-arts demonstrate the superiority of our approach, and the broad application capability, e.g, pose morphing, data augmentation, etc., pushes the margin of domain translation problem.
\end{abstract}
\vspace{-2mm}

\begin{figure}
	\centering
	\includegraphics[scale=0.14]{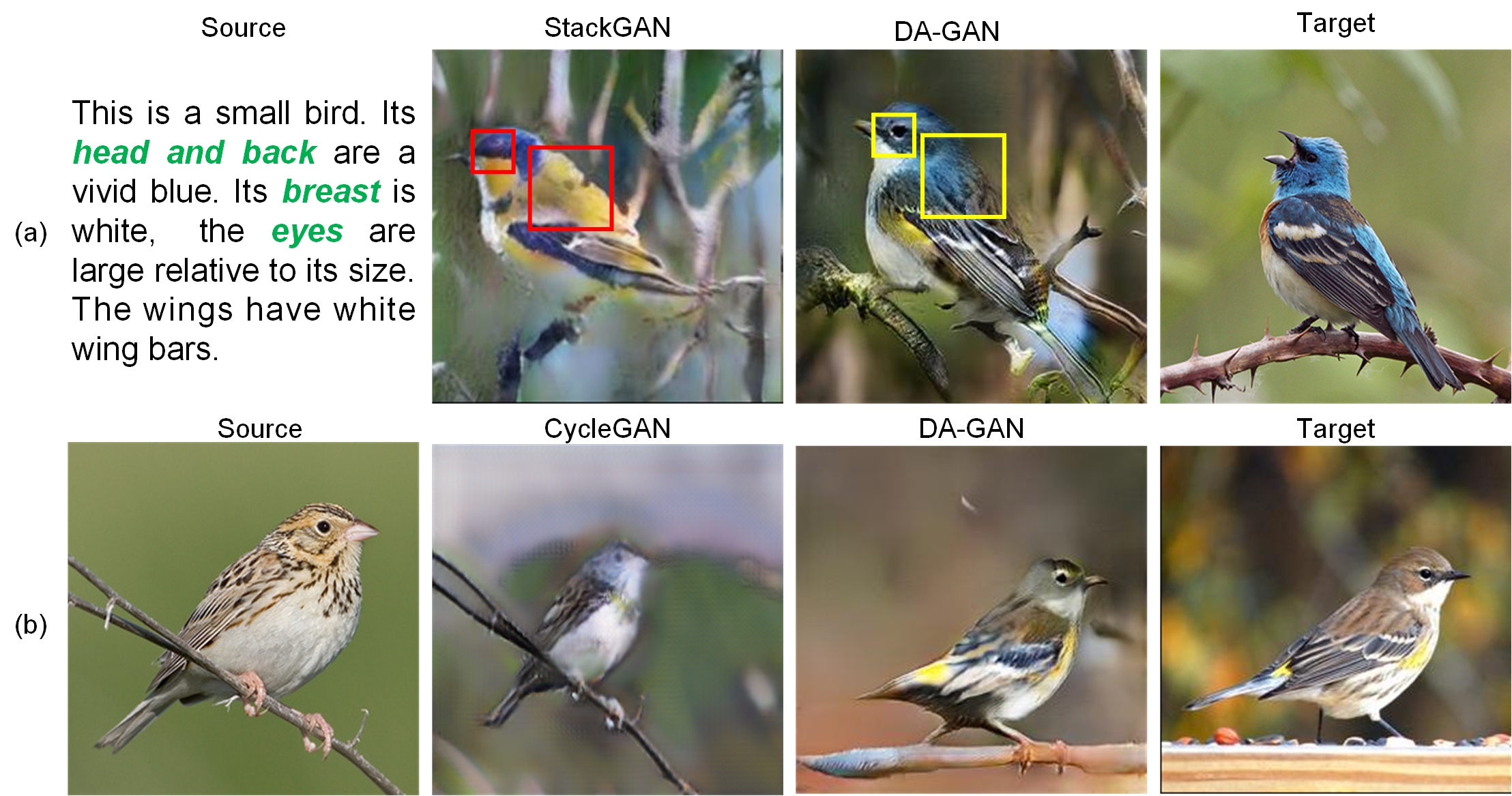}
	\caption{(a) text-to-image generation. (b) object configuration. We can observe that the absence of instance-level correspondences results in both semantic artifacts (labeled by red boxes) exist in StackGAN and geometry artifacts exist in CycleGAN. Our approach successfully produces the correct correspondences (labeled by yellow boxes) because of the proposed instance-level translating. Details can be found in Sec. \ref{intro} }
	\label{title_fig}
	\vspace{-5mm}
\end{figure} 

\vspace{-2mm}
\section{Introduction} \label{intro}
Can machines possess human ability to relate different image domains and translate them? This question can be formulated as image translation problem. In other words, learning a mapping function, by finding some underlying correspondences (e.g. similar semantics), from one image domain to the other. 
Years of research have produced powerful translation systems in supervised setting, where example pairs are available, e.g. \cite{pix2pix2016}. However, obtaining paired training data is difficult and expensive. Therefore, researchers turned to develop unsupervised learning approach which only relies on unpaired data. 
In the unsupervised setting, we only have two independent sets of samples. 
The lacking of pairing relationship makes it considered harder in finding the correct correspondences, and therefore it is much more challenging.
Existing works typically build upon Generative Adversarial Network (GAN) such that the distribution of the translated samples is indistinguishable from the distribution of the target set.
However, we point out that data itself is structured. Such set-level constraint impedes them from finding meaningful instance-level correspondences.
By 'instance-level correspondences', we refer to high-level content involving identifiable objects that shared by a set of samples. These identifiable objects could be adaptively task driven. For example, in Figure \ref{title_fig} (a), the words in the description corresponds to according parts and attributes of the bird image.
Therefore, false positives often occur because of the instance-level correspondences missing in existing works. For example, in object configuration, the results just showing changes of color and texture, while fail in geometry changes (Figure \ref{title_fig}). In text-to-image synthesis, fine-grained details are often missing (Figure \ref{title_fig}). 

Driven by this important issue, a question arises: \textit{Can we seek an algorithm which is capable of finding meaningful correspondences from both set-level and instance-level under unsupervised setting?} To resolve this issue, in this paper, we introduce a dedicated unsupervised domain translation approach builds upon Generative Adversarial Network - \textbf{DA-GAN},  which success in a large variety of translating tasks, and achieve visually appealing results.

To achieve these results, we have to address two fundamental challenges:
First, how to exploit instance-level constraints while lacking correct pairing relationship in unsupervised setting. We take on this challenge and provide the first solution by decomposing the task of translating samples from two independent sets into translating instances in a highly-structured latent space. Specifically, we integrate the attention mechanism into the learning of the mapping function $F$, and a compound loss that consists of a consistency term, a symmetry term and a multi-adversarial term is used. Through attending on meaningful correspondences of samples on instance-level, the learned \textit{Deep Attention Encoder} (DAE) projects samples in a latent space. Then the constraint on instance-level could be exploited in the latent space. We introduce a \textit{consistency loss} to require the translated samples correspond to correct semantics with samples from the source domain in the latent space. To further enhance the constraint, we also consider the samples from the target domain by adding a \textit{symmetry loss} that encourages the one-to-one mapping of $F$. As a result, the instance-level constraints enable the mapping function to find the meaningful semantic corresponding, and therefore producing true positives and visually appealing results.

Second, how to further strengthen the constraints on set level such that the mode collapse problem could be mitigate.
In practical, all input samples will map to the same sample, and optimization fails to make progress. 
To address this issue, we introduce a multi-adversarial training procedure to encourage different modes achieve fair possibility mass distribution during training and thus providing an effective solution to encourage the mapping function could cover all modes in the target domain, and make progress to achieve the optimal.
Our main contributions can be summarized into three-fold:
\vspace{-2mm}
\begin{itemize}
	\setlength{\itemsep}{-3pt}
	\item We decompose the task to instance-level image translation such that the constraints could be exploited on both instance-level and set-level by adopting the proposed compound loss.
	\item To the best of our knowledge, we are the first that integrate the attention mechanism into Generative Adversarial Network.
	\item We introduce a novel framework DA-GAN, which produces visually appealing results and is applicable in a large variety of tasks.
\end{itemize}
\vspace{-2mm}

\begin{figure*}
	\centering
	\includegraphics[scale=0.25]{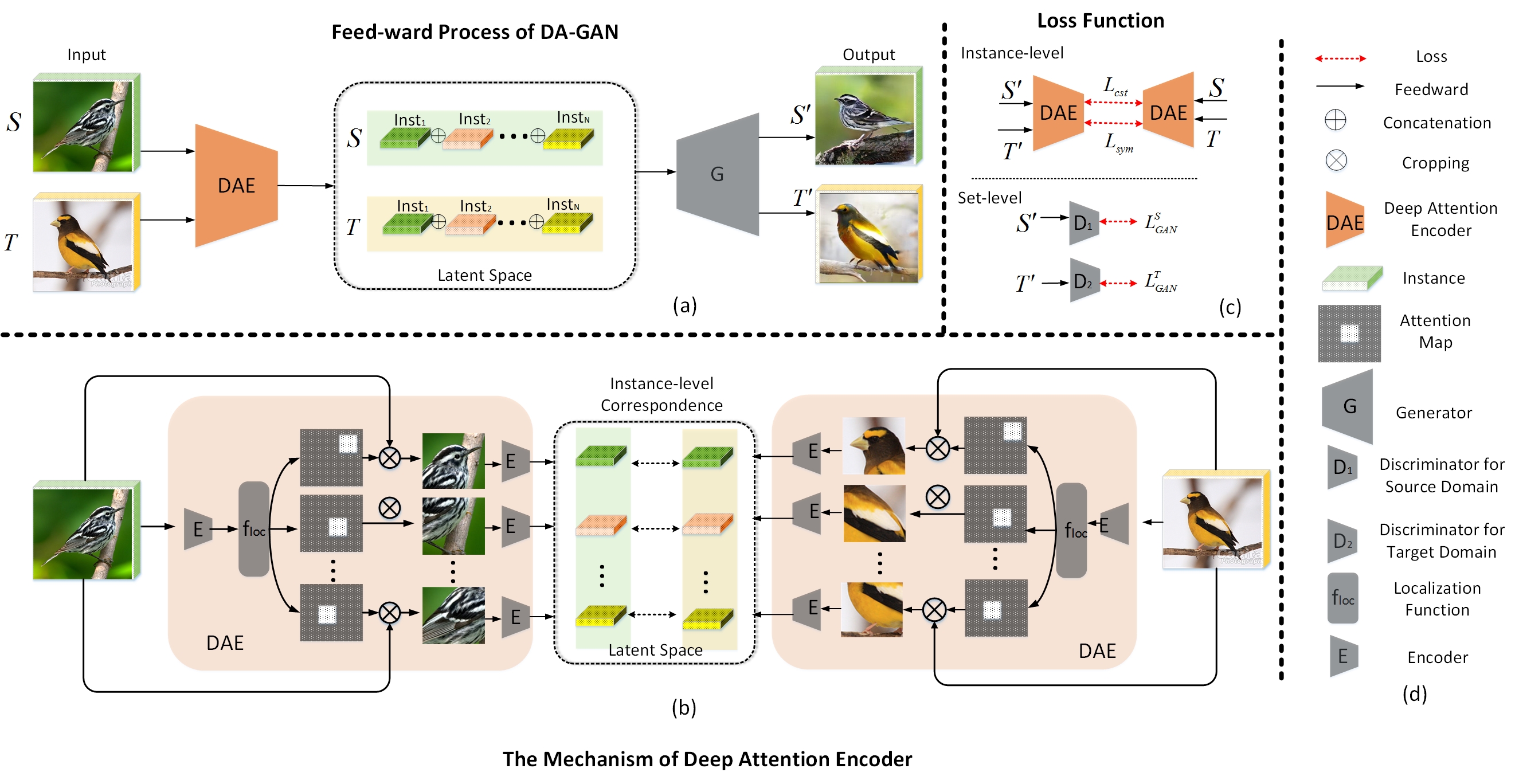}
	\caption{A pose morphing example for illustration the pipeline of DA-GAN. Given two images of birds from source domain $S$ and target domain $T$, the goal of pose morphing is to translate the pose of source bird $s$ into the pose of target one $t$, while still remain the identity of $s$. The feed-ward process is shown in (a), where two input images are fed into DAE which projects them into a latent space (labeled by dashed box). Then G takes these highly-structured representations ($DAE(s)$ and $DAE(t)$) from the latent space to generated the translated samples, i.e.$s' = G(DAE(s))$, $t' = G(DAE(t))$. The details of the proposed DAE (labeled by orange block) is shown in (b). Given an image $X$, a localization function $f_{loc}$ will first predict $N$ attention regions' coordinates from the feature map of $X$, (i.e. $E(X)$, where $E$ is an encoder, which can be utilized in any form). Then $N$ attention masks are generated and activated on $X$ to produce $N$ attention regions $\{ {R_i}\} _{i = 1}^N$. Finally, each region's feature consists the instance-level representations $\{ Ins{t_i}\} _{i = 1}^N$. By operating the same way on both $S$ and $T$, the instance-level correspondences can consequently be found in the latent space. We exploit constraints on both instance-level and set-level for optimization, it is illustrated in (c). All of the notations are listed in (d). [Best viewed in color.]}
	\label{pipeline}
	\vspace{-2mm}
\end{figure*} 
\section{Related Work}
\subsection*{Generative Adversarial Networks}
Since the Generative Adversarial Networks (GANs) was proposed by Goodfellow
et al., \cite{GoodFellow:GAN:NIPS2014} researchers have studied it vigorously.
Several techniques have been proposed to stabilize the training techniques \cite{DBLP:Alec:GAN, DBLP:MetzPPS16, DBLP:journals/corr/SalimansGZCRC16, DBLP:conf/icml/ArjovskyCB17:WGAN, DBLP:journals/corr/ZhaoML16} and generate compelling results. 
Built upon these generative models, several methods were developed to generate images based on GAN. Most methods utilized conditioning variables such as attributes or class labels \cite{DBLP:journals/corr/YanYSL15, DBLP:journals/corr/OordKVEGK16, DBLP:journals/corr/ChenDHSSA16:InfoGAN, pmlr-v70-odena17a, DBLP:journals/corr/MirzaO14:conditionalGAN}. There are also works conditioned on images to generate images, e.g. photo editing \cite{DBLP:journals/corr/BrockLRW16a, DBLP:journals/corr/HeZRS15}, and super-resolution \cite{DBLP:journals/corr/LedigTHCATTWS16, DBLP:journals/corr/SonderbyCTSH16}. 
Other approaches used conditional features from a completely different
domain for image generation. Reed et al. \cite{Reed:2016:ICML} used encoded text description of images as the conditional information to generating 64 $\times$ 64 images that match the description. Their follow-up work \cite{Reed:2016:ICML} can produce 128 $\times$ 128 images by utilizing additional annotations on object part locations. In StackGAN \cite{Zhang:16:StackGAN}, two GANs in different stages are adopted to generate high resolution images. Comparing with StackGAN, the proposed DA-GAN can generated 256 $\times$ 256 images directly. More importantly, we trained the network by unpaired data, and achieve visually appealing results.

\subsection*{Image-to-Image Translation}
"pix2pix"\cite{pix2pix2016} of Isola et al., which uses a conditional GAN \cite{GoodFellow:GAN:NIPS2014} to learn a mapping from input to output images. Similar ideas have been applied to various tasks such as generating photographs from sketches \cite{DBLP:journals/corr/SangkloyLFYH16} or from
attribute and semantic layouts \cite{DBLP:journals/corr/KaracanAEE16}. 
Recently, \cite{Ganin:2016:DTN} proposed the domain transformation network (DTN) and achieved promising results on translating small resolution face and digit images. CoGAN \cite{Ming:NIPS:2017:Nvidia} and cross modal scene networks \cite{DBLP:journals/corr/AytarCVPT16} use a weight-sharing strategy to learn a common representation across domains. 
Another line of concurrent work \cite{DBLP:journals/corr/BousmalisSDEK16, DBLP:journals/corr/ShrivastavaPTSW16, Yaniv:ICLR:2017:emoji} encourages the input and output to share certain content features even though they may differ in style. They also use adversarial networks, with additional terms to enforce the output to be close to the input in a predefined metric space, such as class label space, image pixel space, and image feature space.
In CycleGAN \cite{Jun:ICCV:2017:CycleGAN}, a cycle consistency loss is proposed to enforce one-to-one mapping. We note that several contemporary works \cite{Yi:ICCV:2017:DualGAN, Taeksoo:ICML:2017:DiscoGAN} are all introduced the cycle-consistency constraint for the unsupervised image translation.
Neural Style Transfer \cite{Gatys:StyleTransfer:CVPR2016} \cite{JohnsonAL16:RealTimeStyleTransfer, UlyanovLVL16} is another way
to perform image-to-image translation, which synthesizes a
image by combining the content of one image with
the style of another image based on pre-trained deep features. 
Different with style transfer, domain translation aims in learning the mapping between two image collections, rather than between two specific images. 

\vspace{-2mm}
\section{Approach}
Our aim is to learn a mapping function $F$ that maps samples from source domain $S:\{ {s_i}\} _{i = 1}^N$ to target domain $T:\{ {t_i}\} _{i = 1}^M$, denoted as $F:S \to T$. As illustrated in Figure \ref{pipeline}, the proposed DA-GAN consists of four modules: a Deep Attention Encoder ($DAE$), a Generator($G$) and two discriminators ($D_1$, $D_2$). The mapping is conducted from both source domain and target domain. The translated samples sets from source domain and target domain are denoted as $S'$ and $T'$, respectively. We introduce the $DAE$ in Sec. \ref{attention}. The translation on instance-level and set-level are introduced in Sec. \ref{intance_level} and in Sec. \ref{set_level}, respectively.

\vspace{-2mm}
\subsection{Deep Attention Encoder} \label{attention}
To project samples into the latent space, we integrate attention mechanism to jointly learn an Deep Attention Encoder $DAE$.
Given a feature map $E(X)$ of an input image $X$ (where $E$ is an encoder that could be utilized in any form), we first adopt a localization function ${f_{loc}}( \cdot )$ to predict a set of attention regions' location, which is given by:
\vspace{-2mm}
\begin{equation}
{f_{loc}}(E(X)) = [{x_i},{y_i}]_{i = 1}^{N'},
\vspace{-2mm}
\end{equation}
where $[{x_i},{y_i}]$ denotes a region's center coordinates, $N'$ denotes the number of regions predicted. Once the the location of an attended region is hypothesized, we generate an attention mask $\mathcal{M}_i$.
Specifically, we denote $w$ and $h$ as half of the width and half of the height of $X$. Then we can adopt the parameterizations of attend region by:
\begin{equation}
\begin{array}{l}
\begin{array}{l}
x_i^{left} = {x_i} - w,\;\;\;x_i^{right} = {x_i} + w,\\
y_i^{top} = {y_i} - h,\;\;\;y_i^{bottom} = {y_i} + h.
\end{array}
\end{array}
\end{equation}
The cropping operation can therefore be achieved by an element-wise multiplication applied on $X$, i.e. ${R_i} = X^\circ {{\cal M}_i}$, which produces the attended regions  $\{ R_i\} _{i = 1}^{N'}$. Then instance-level representations of $X$ in the latent space are defined by:
\vspace{-2mm}
\begin{equation}
\{ E({R_i})\} _{i = 1}^{N'} = \{ Inst\} _{i = 1}^{N'},
\vspace{-2mm}
\end{equation}
To allow backpropagation, here we adopt the attention mask as:
\vspace{-2mm}
\begin{equation}
\begin{array}{l}
{{\cal M}_i} = [\sigma (x - x_i^{left}) - \sigma (x - x_i^{right})] \cdot \\
\;\;\;\;\;\;\;\;\;\;[\sigma (y - y_i^{top}) - \sigma (y - y_i^{bottom})],
\end{array}
\end{equation}
where $\sigma (\cdot) = 1/(1 + {\exp ^{ - kx}})$ is a sigmoid function.
In theory, when $k$ is large enough, $\sigma(\cdot)$ is approximated as a step function and $\mathcal{M}_i$ will become a two dimensional rectangular function, then the derivation could be approximated.
For learning these attention regions, we add a geometric regularization ${\mathbb{E}_{X \sim {P_{data}}(X)}}[d({Y},DAE(X))]$. $Y$ is the label of image $X$, and $d$ is some similarity metrics in the data space,  In practice, there are many options for the distance measure $d$. For instance, a VGG classifier.

\subsection{Instance-Level Image Translation} \label{intance_level}
As the DAE projects $s$ and $t$ into a shared latent space, we can constrain them to be matched with each other in this latent space. Therefore, we adopt a consistency loss on the samples from source domain $\{ {s_i}\} _{i = 1}^N$ and the according translated samples $\{ {{s'}_i}\} _{i = 1}^N$:
\vspace{-2mm}
\begin{equation}
\mathcal{L}_{cst} = {\mathbb{E}_{s \sim {P_{data}}(s)}}\;d(DAE(s),\;DAE(F(s)),
\vspace{-2mm}
\end{equation}

On the other hand, we also consider the samples from the target domain to further enforce the mapping to be deterministic. In theory, if a mapping is bijective (one-to-one corresponding), the operation from a set to itself form a symmetric group. The mapping can then be considered as a permutation operation on itself. We therefore exploit a symmetry loss to enforce $F$ can map samples from $T$ to themselves, i.e. ${t_i} \approx F({t_i})$. The loss function is defined as:
\vspace{-2mm}
\begin{equation}
{\mathcal{L}_{sym}} = {\mathbb{E}_{t \sim {P_{data}}(t)}}\;d(DAE(t),\;DAE(F(t)),
\vspace{-2mm}
\end{equation}
this can also be considered as an auto-encoder type of loss applied on samples from $T$, where $d$ is a distance measure. In theory, there are many options for $d$. For instance, the $L^n$ distance, or the distance of learned features by the discriminator or by other networks, such as a VGG classifier.

\subsection{Set-Level Image Translation} \label{set_level}
It is straight-forward to use a discriminator $D_1$ to distinguish the translated samples $\{ {{s'}_i}\} _{i = 1}^N$ from the real samples in the target domain $\{ t\} _{i = 1}^M$, and generator is forced to translate samples that is indistinguishable from real samples in target domain, which is given by:
\vspace{-2mm}
\begin{equation}
\begin{array}{l}
{\mathcal{L}_{GAN}^s} = {\mathbb{E}_{t \sim {P_{data}}(t)}}[\log {D_1}(t)]\\
\;\;\;\;\;\;\;\;\;\;\; + \;{\mathbb{E}_{t \sim {P_{data}}(s)}}[\log (1 - {D_1}(F(s)))].
\end{array}
\vspace{-2mm}
\end{equation}
While there still exists another issue - mode collapse. In theory, large modes usually have a much higher chance of attracting the gradient of discriminator, and the generator is not penalized for missing modes. In practice, all input samples map to the same output, and the optimization fails to make progress. This issue asks for adding penalty on generator for missing modes.

As we mentioned before, $ DAE\circ G$ can be considered as an auto-encoder for 
$\{ t_i\} _{i = 1}^M$. Then for every modes in $T$, $F(t)$ is expected to generate very closely located modes. 
We therefore add another discriminator $D_2$ for samples from the target domain to enforce the reconstructed $t'$ is indistinguishable from $t$. An additional optimization objective for the generator is hence added
${\mathbb{E}_{t \sim {p_{data}(t)}}}[\log D_2(F(t))]$. The objective function is given by:
\vspace{-2mm}
\begin{equation}
\begin{array}{l}
{\mathcal{L}_{GAN}^t} = {\mathbb{E}_{t \sim {P_{data}}(t)}}[\log {D_2}(t)]\\
\;\;\;\;\;\;\;\;\;\;\; + \;{\mathbb{E}_{t \sim {P_{data}}(t)}}[\log (1 - {D_2}(F(t)))].
\end{array}
\vspace{-2mm}
\end{equation}
This multi-adversarial training procedure is critical for penalizing the missing modes, it encourage $F(t)$ to move towards a nearby mode of the data generating distribution. In this way, we can achieve fair probability mass distribution across different modes. 
\begin{figure}
	\centering
	\includegraphics[width=\linewidth]{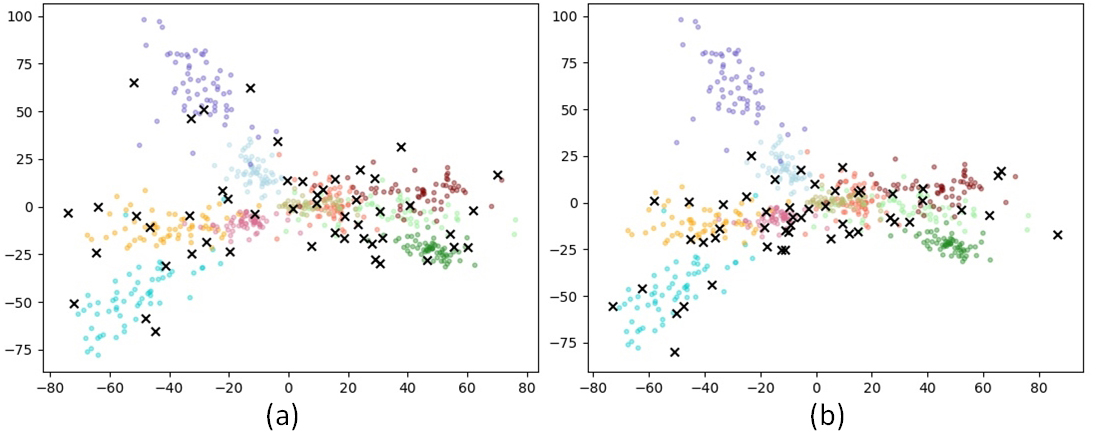}
	\caption{Visualized distribution of 10 classes of birds. Each color represents a birds class. Black crosses represents the distribution of the generated samples. (a): generated data distribution of DA-GAN. (b) generated data distribution of StackGAN \cite{Zhang:16:StackGAN}.}
	\label{text2img_distribution}
	\vspace{-3mm}
\end{figure} 
 
 \begin{figure}[t]
	\centering
	\includegraphics[scale=0.46]{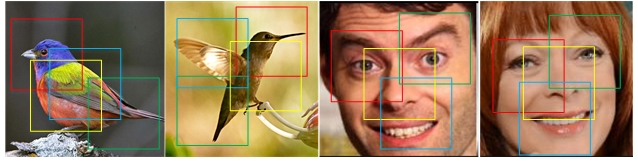}
	\caption{The attention locations predicted by DAE on birds images and face images from.}
	\label{encoder_fig}
\end{figure} 
\subsection{Full Objective and Implementation Details} \label{Implementation}
Our full objective is given by:
\vspace{-2mm}
\begin{equation}
\begin{array}{l}
\mathcal{L}(DAE,G,{D_1},{D_2}) = \mathcal{L}_{GAN}^s(DAE,G,{D_1},S,T)\\
\;\;\;\;\;\;\;\;\;\;\;\;\;\;\;\;\;\;\;\;\;\;\;\;\;\;\;\;\;\;\;\;+ \mathcal{L}_{GAN}^t(DAE,G,{D_{2,}}T)\\
\;\;\;\;\;\;\;\;\;\;\;\;\;\;\;\;\;\;\;\;\;\;\;\;\;\;\;\;\;\;\;\; + \alpha {\mathcal{L}_{cst}}(DAE,G,S)\\
\;\;\;\;\;\;\;\;\;\;\;\;\;\;\;\;\;\;\;\;\;\;\;\;\;\;\;\;\;\;\;\; + \beta {\mathcal{L}_{sys}}(DAE,G,T),
\end{array}
\vspace{-2mm}
\end{equation}
where $\alpha$ and $\beta$ are weights for the consistency loss and symmetry loss, respectively. We aim to solve:
\vspace{-2mm}
\begin{equation}
{F^ * } = \arg \mathop {\min }\limits_F \mathop {\max }\limits_{{D_1},{D_2}} \mathcal{L}(F,{D_1},{D_2})
\vspace{-2mm}
\end{equation}
where $F = DAE \circ G$.

We adopt the generator consists of several residual blocks \cite{DBLP:journals/corr/HeZRS15}. For the generator, the instance-level representations are concatenated along the channel dimension and fed into several residual blocks. Finally, a series of up-sampling layers are used to generate a the translated image. 
For the discriminator, the generated image is fed through a series of down-sampling blocks. Finally, a fully-connected layer with one node is used to produce the decision score.
The up-sampling blocks consist of the nearest-neighbor
upsampling followed by a 3$\times$3 stride 1 convolution. Batch
normalization and ReLU activation are applied after
every convolution except the last one. The residual blocks
consist of 3$\times$3 stride 1 convolutions, Batch normalization
and ReLU. 
All networks are trained using Adam solver with batch size 64 and an initial learning rate of 0.0002.
\vspace{-2mm}
\begin{figure*}[h!]
	\centering
	\includegraphics[scale=0.28]{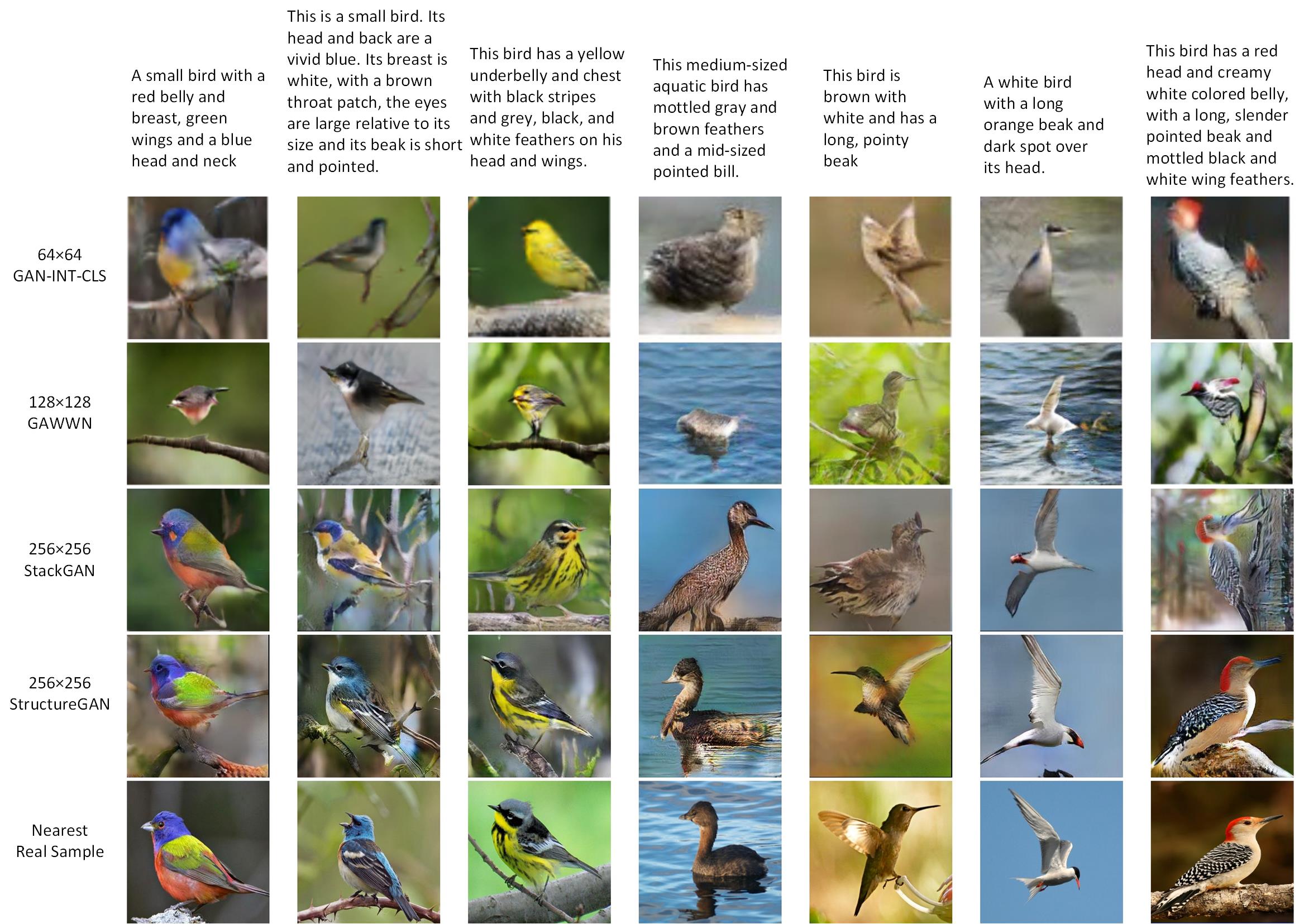}
	\caption{Examples results of text to image synthesis.}
	\label{text2img_fig}
\end{figure*} 

\begin{figure*}
	\centering
	\includegraphics[scale=0.36]{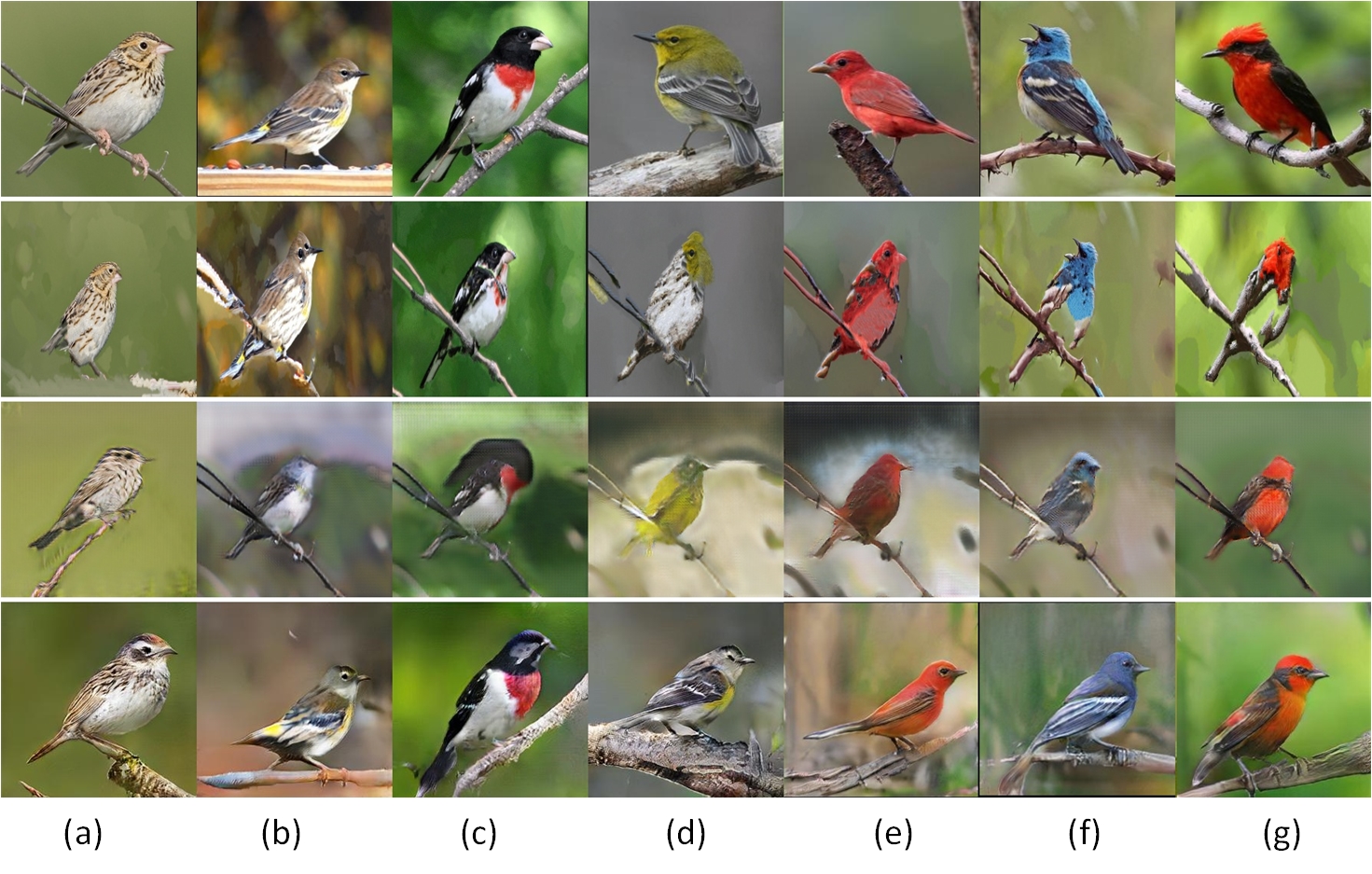}
	\vspace{-2mm}
	\caption{Example results of object configuration. Each row from top to bottom are the real samples, results generated by VAT\cite{Liao:2017:VisualAttribute}, CycleGAN \cite{Jun:ICCV:2017:CycleGAN} and DA-GAN, respectively.}
	\label{configuration}

\end{figure*}

\begin{figure*}
	\centering
	\includegraphics[scale = 0.2]{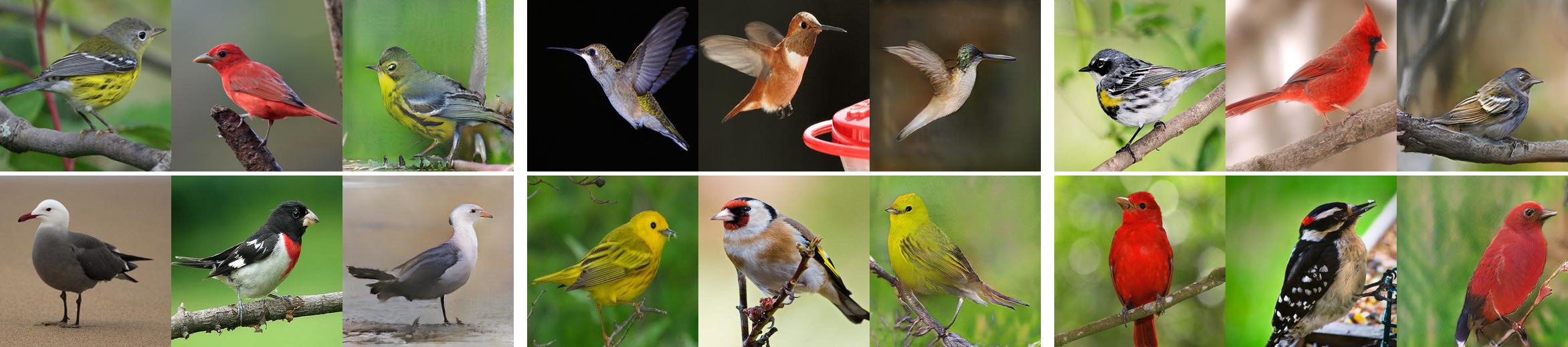}
	\caption{Example results of pose morphing. In each group, the first column are a source bird $s$, the second column are the target bird $t$, the third column are birds that generated by DA-GAN}
	\label{pose}
	\vspace{-3mm}
\end{figure*}

\begin{figure*}[t]
	\centering
	\includegraphics[scale=0.45]{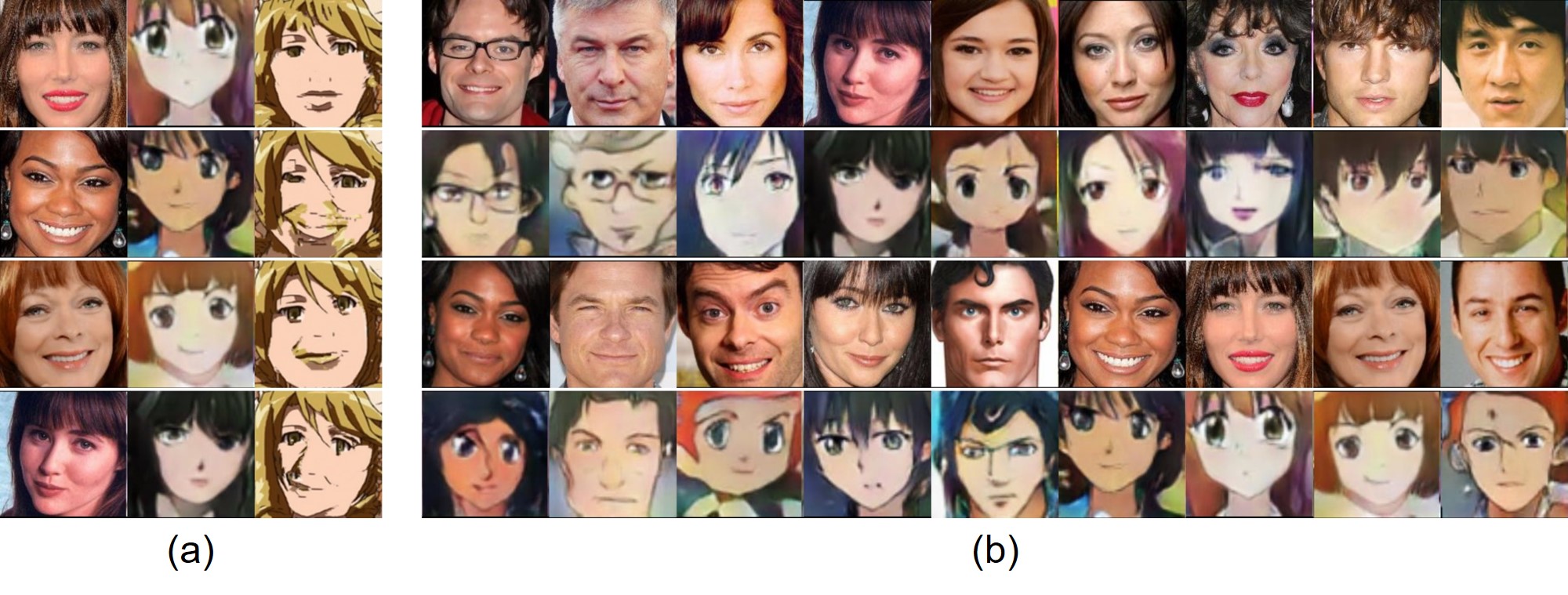}
	\vspace{-3mm}
	\caption{(a) Comparisons of face translation with VAT. Each row from left to right is human face, results produced by DA-GAN and VAT, respectively. (b) Human face to animation face synthesis. Human faces are placed in the first and third rows, the according translated animation faces are placed in the second row and fourth row.}
	\label{face}

\end{figure*}

\begin{figure*}
	\centering
	\includegraphics[scale=0.28]{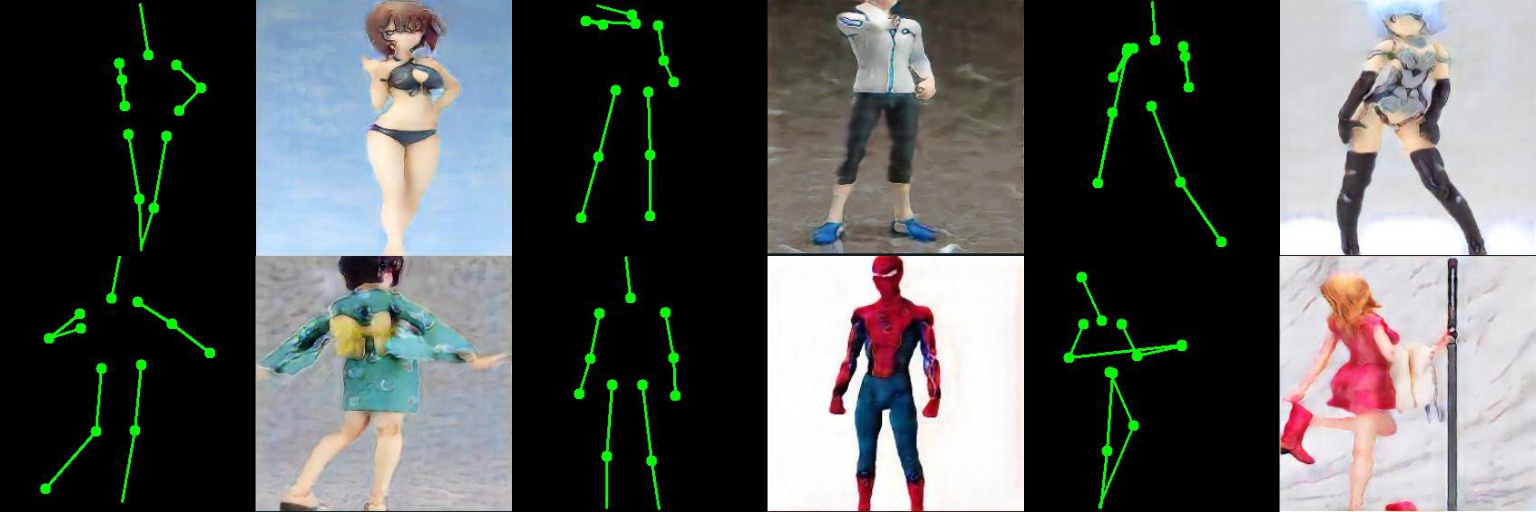}
	\caption{Example results of skeleton to cartoon figure synthesis.}
	\label{skeleton}
	\vspace{-1mm}
\end{figure*}


\section{Experiments}
In this section, we validate the effectiveness of the proposed DA-GAN in a large variety of tasks, including domain adaption, text-to-image synthesis, object configuration, pose morphing for data augmentation, face-to-animation synthesis and skeleton to cartoon figure synthesis. 
We conduct these experiments on several datasets, including MNIST \cite{lecun:2010:MNIST},  CUB-200-2011 \cite{WahCUB_200_2011}, SVHN \cite{DBLP:SVHN}, FaceScrub \cite{ng2014a} and AnimePlanet \footnote{It is retrieved from http://www.anime-planet.com/, which has about 60k images.}. 

\vspace{-1mm}
\subsection{Baselines}
 \begin{itemize}
	\setlength{\itemsep}{-3pt}	
	\item \textbf{GAN-INT-CLS}  \cite{Reed:2016:ICML} succeeds in synthesizing 64 $\times$ 64 birds and flowers images based on text descriptions.
	\item \textbf{GAWWN} is Reed's follow-up work \cite{Reed:NIPS:2016} that was able to generate 128 $\times$ 128 images.
	\item \textbf{StackGAN} is the latest work that can synthesize high-quality images in 256 $\times$ 256, from text descriptions. 
	\item \textbf{SA} is an early work that explored ideas from subspace learning for domain adaption \cite{Fernando:2013:ICCV:DomainAdaption}. 
	\item \textbf{DANN} It is another domain adaption work that conducted by  \cite{Ganin:2016:DTN} deep feature learning.
	\item \textbf{UNIT} is a recent unsupervised image-to-image translation work \cite{Ming:NIPS:2017:Nvidia} which based on the shared-latent space assumption and cycle loss. 
	\item \textbf{DTN} \cite{Yaniv:ICLR:2017:emoji} employs a compound loss function for unsupervised domain translation. 
	\item \textbf{CycleGAN} is an image-to-image translation work that adopt GAN with cycle-loss \cite{Jun:ICCV:2017:CycleGAN, Taeksoo:ICML:2017:DiscoGAN, Ming:NIPS:2017:Nvidia,Yi:ICCV:2017:DualGAN}. 
	\item \textbf{VAT} \cite{Liao:2017:VisualAttribute} is a new technique derives from style transfer, while it different in finding dense correspondences. 
\end{itemize}
\vspace{-2mm}

\vspace{-2mm}
\subsection{Component Analysis of DA-GAN} \label{component_analysis}
We trained a classifier on MNIST dataset and employ it on the translated samples for quantitative evaluation.
The results are shown in Table \ref{multi-table}. As we can see, the DA-GAN approaches very high accuracy on the translated sample set. While the results is impaired without the DAE. We also fine-tune a VGG \cite{simonyan2015very} classifier on on the CUB-200-2011, and use it to test our generated images from text, The accuracy drops a lot to 60.6 $\%$. 
We also show some results produced by DAE in Figure \ref{encoder_fig}. It can be seen that $f$ is capable of attending on semantic regions. For example, birds head, wings, and etc. human's eyes, mouth, and etc.

\begin{table*}
\begin{minipage}[b]{.3\linewidth}
\centering

\begin{tabular}{c|c} 
	\hline
	Method & Accuracy \\
	\hline
	DA-GAN & 94.3 $\%$ \\
	DA-GAN w/o $DAE$ & 90.2 $\%$ \\
	DA-GAN w/o const & 79.8$\% $ \\
	DA-GAN w/o Sym & 90.6$\% $ \\
	DA-GAN w/o $D_2$ & 88.2$ \%$ \\
	\hline
\end{tabular} 

\subcaption{}

\end{minipage}
\begin{minipage}[b]{.3\linewidth}
\centering

\begin{tabular}{c|c} 
	\hline
	Method & Accuracy \\
	\hline
	SA\cite{Fernando:2013:ICCV:DomainAdaption} & 59.32 $\%$ \\
	\hline
	DANN\cite{Ganin:2016:DTN} & 73.85 $\%$ \\
	\hline
	DTN \cite{Yaniv:ICLR:2017:emoji} & 84.44$\%$ \\
	\hline
	UNIT \cite{Ming:NIPS:2017:Nvidia} & 90.53 $\%$ \\
	\hline
	DA-GAN & \textbf{93.60} $\%$ \\
	\hline
\end{tabular} 
\subcaption{}

\end{minipage}
\begin{minipage}{.4\linewidth}
	\centering
	\vspace{-27mm}
	\begin{tabular}{c|c|c} 
		\hline
		Method & Inception & $\sharp$ miss\\
		\hline
		GAN-INT-CLS & 2.9 $ \pm $ 0.4 & 89.0 \\
		GAWWN & 3.6 $ \pm $ 0.4 & 61.0\\
		StackGAN & 3.7 $ \pm $ 0.4 & 36.0\\
		DA-GAN & \textbf{5.6$ \pm $ 0.4}  & \textbf{19.0}\\
		\hline
	\end{tabular}  
\subcaption{}

\end{minipage}
\caption{(a): Component evaluation of DA-GAN. (b): Comparisons with state-of-the-arts on domain adaption. (c): Comparisons with state-of-the-arts on text-to-image synthesis.}
\vspace{-3mm}
\label{multi-table}
\end{table*}

To validate that the proposed DA-GAN is effective in mitigating the mode collapse problem. We conduct a toy experiment on a subset of samples from CUB-200-2011. We select 10 classes of birds. To mimic the large mode, we picked some similar classes (e.g. some of them are from the same category). The dense region in Figure. \ref{text2img_distribution} shows the birds that have similar looking. We generate about 600 images by input the according text descriptions and the distribution of the generated data is shown in Figure \ref{text2img_distribution}(a). The same setting is conducted on StackGAN, the results is shown in \ref{text2img_distribution}(b). As we can see that, comparing with StackGAN, the samples generated by DA-GAN are more divers, and larger coverage.

\subsection {Domain Adaptation} \label{domainAdaption_sec}


We applied the proposed framework to the problem of domain adaption, i.e. adapting a classifier trained using labeled samples in one domain (source domain) to classify samples in a new domain (target domain) where labeled samples in the new domain are unavailable during training. For this purposes, we transform images from SHVN to the MNIST domain. The results of this experiment are reported in Table \ref{multi-table}. We found that our method achieved a 94.6 $\%$ accuracy for the SVHN to MNIST translation task, which was much better than 90.53 $\%$ achieved by the previous state-of-the-art method.

\subsection{Text to Image Synthesis} \label{text2img_sec}
We conduct qualitative and quantitative evaluation on the text-to-image synthesis task. Comparisons with several state-of-the-arts \cite{Zhang:16:StackGAN, Reed:NIPS:2016, Reed:2016:ICML} on CUB-200-2011 dataset are shown in Figure \ref{configuration}. 
The quantitative evaluation are measured by two metrics: inception score \cite{DBLP:Inception} and the number of missing modes (denote as $\sharp$ miss).
The inception score is defined as:
\begin{equation}
I = \exp ({E_x}{D_{KL}}(p(y|x)||p(y))),\;\;\;
\end{equation}
where $x$ denotes one generated sample, and $y$ is the label predicted by the Inception model. 
In our experiments, we fine-tune a VGG19 model which is introduced in Sec. \ref{component_analysis}. While the inception score is considered as a good assessment for sample quality. However, the inception score is sometimes not a good metric for missing modes evaluation. 
For stronger validation, we adopt another evaluation metric - missing mode ($\sharp$miss) It represents the classifier reported number of missing modes, i.e. the size of the numbers that the model never generates.
As shown in Table \ref{multi-table}(c), DA-GAN achieves much improvements in terms of inception score, and the missing modes drop dramatically,  which again proves the effectiveness of our proposed framework.
Some examples results are shown in Figure \ref{text2img_fig} for a visualized comparison.

\subsection{Object Transfiguration } \label{configure_section}
We use images of seven classes form the CUB-200-2011 dataset to perform object configuration, i.e. translate a source bird into a target breed. 
Some example results are show in Figure \ref{configuration}. 
The first row is real samples form each breed, and we aims in translating bird (a) into the following six breeds. Among these selected target birds, (b) is selected as the most similar one with (a) in both spatial and geometry attributes. (c) is selected sharing similar spatial attribute while different in geometry attribute. (d-e) are all selected that have different spatial and geometry attributes with (a). 
We can see that, without similar semantic structure, VAT \cite{Liao:2017:VisualAttribute} fails in translating birds, due to their limited corresponding matching method. CycleGAN \cite{Jun:ICCV:2017:CycleGAN} is robust to spatial changes while fail in changing the birds geometries. Comparing with the results that produced by DA-GAN, both shows blurred images that missing fine-grained details.
We can see that, DA-GAN succeeds in translating images that even have large variance of spatial and geometry attributes. It strongly validates our claim that the instance-level corresponding is critical in translation task.
We further conducted quantitative evaluation, which can be found in Table \ref{config_table}. The images produced by DA-GAN out performs in both classification accuracy and realism.

\begin{table} 
	\begin{center}	
		\begin{tabular}{c|c|c|c|c} 
			\hline
			& Real & VAT & CycleGAN & DA-GAN \\
			\hline\hline
			Top-1 acc & 98.6 $\%$ & 42.1 $\%$ & 62.1 $\%$ & \textbf{88.9 $\%$} \\
			Realism & 20.9 & 10.3 & 11.4 & \textbf{18.9} \\
			\hline
		\end{tabular} 
	\end{center} 
	\vspace{-3mm}
	\caption{Qualitative evaluations for object configuration.}
	\label{config_table} 
	\vspace{-4mm}
\end{table}

\subsection{More Applications} \label{pose_sec}
We further conduct pose morphing, which considered harder in changing the geometries, by DA-GAN. The results are shown in Figure \ref{pose}. It can be seen that, we succeed in morphing the birds' pose even when there exists very large gap of geometry variance.
For practical usage, we also make used of these morphed samples for data augmentation. For each image, we randomly picked 10 references as the pose targets. Top-1 result is picked for each image and is used for augmented data, which produced about 10K images of birds. We then applied a pre-trained VGG on the augmented data, which shows improvement on fine-grained classification task. The results is shown in Table \ref*{dataaug}. 
\begin{table} 
	\begin{center}	
		\begin{tabular}{c|c|c} 
			\hline
			Method & Training Data & Accuracy \\
			\hline\hline
			no data augmentation & 8K & \textbf{79.0} $\%$ \\
			DA-GAN & 8K + 10K & \textbf{81.6} $\% $ \\
			\hline
		\end{tabular} 
	\end{center} 
	\vspace{-3mm}
	\caption{Data augmentation results.}
	\label{dataaug} 
	\vspace{-6mm}
\end{table}

We adopt the DA-GAN to translate a human face into a animation face while still preserve the human identity, the results are shown in Figure \ref{face}.
We also compare our results with the ones produces by VAT \cite{Liao:2017:VisualAttribute} in Figure \ref{face}(a). We can see that, VAT cannot solve the task we are tackling. The produced images does not belong to the target domain, i.e. an animation face. More severely, when two query face shows different shooting angle, VAT produces artifacts due to the incorrect semantic correspondences. 
More experiment are conducted on skeleton to cartoon figure translation. The results are shown in Figure \ref{skeleton}. More experimental details can be found in supplementary materials.

\section{Conclusion}
In this paper, we propose a novel framework for unsupervised image translation. Our intuition is to decompose the task of translating samples from two sets into translating instances in a highly-structured latent space. 
The instance-level corresponding could then be found by integrating attention mechanism into GAN. Extensive quantitative and qualitative results validate that, the proposed DA-GAN can significantly improve the state-of-the-arts for image-to-image translation.
It is superiority in scalable for broader application, and succeeds in generating visually appealing images. 
We find that, some failure cases are caused by the incorrect attention results. It is because the instances are learned by a weak supervised attention mechanism, which some time showing a large gap with that learned under fully supervision. To tackle this challenge we may seek for more robust and effective algorithm in the future.


\clearpage
{\small
\bibliographystyle{ieee}
\bibliography{egbib}
}

\clearpage

\subsubsection*{Implementation Details}
\vspace{-2mm}
The experimental settings for each task are listed in Table \ref{detail}. 
'$\sharp$' denotes the number of attention regions that are pre-defined in each task, and 'Instances' denotes the attended level of the instances. The label $Y$ and the distance metric $d(\cdot)$ are adopted in the optimization of Deep Attention Encoder (DAE) and the instance-level translation. Note that $d$ is jointly trained from scratch with DA-GAN, where 'ResBlock' denotes a small classifier that consists of 9 residual blocks. The learned attention regions are adaptively controlled by the selection of $Y$, $\sharp$ and $d(\cdot)$. 
For example, the instances we learned on tasks conducted on CUB-200-2011 are parts level (birds' four parts), and for task of Colorization and domain adaption, the attended instances are objects (flower and characters). 
\vspace{-3mm}

\subsubsection*{Experiments on CUB-200-2011}
\begin{figure}
	\centering
	\includegraphics[scale=0.7]{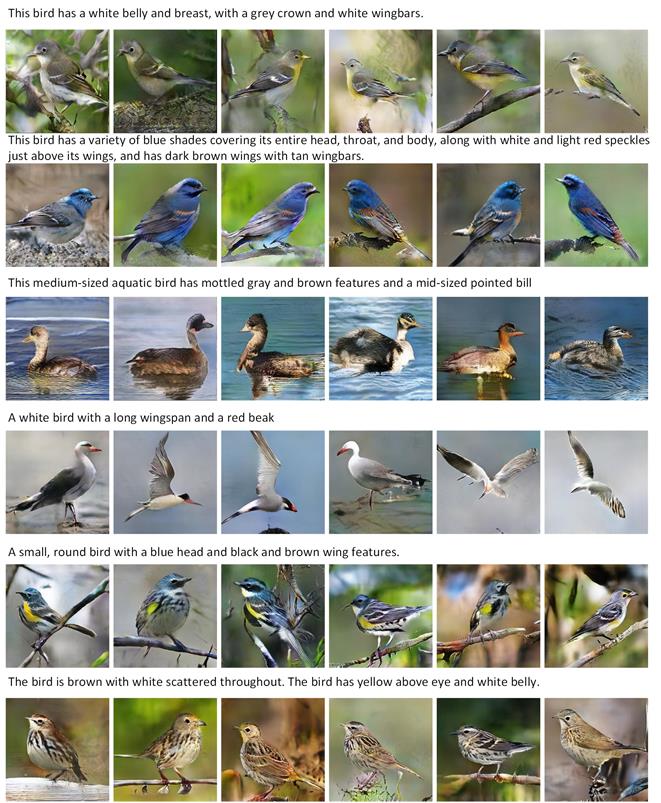}
	\caption{Experimental Results of text-to-image synthesis.}
	\label{text2img}
\end{figure}

\begin{figure*}
	\centering
	\includegraphics[scale=0.7]{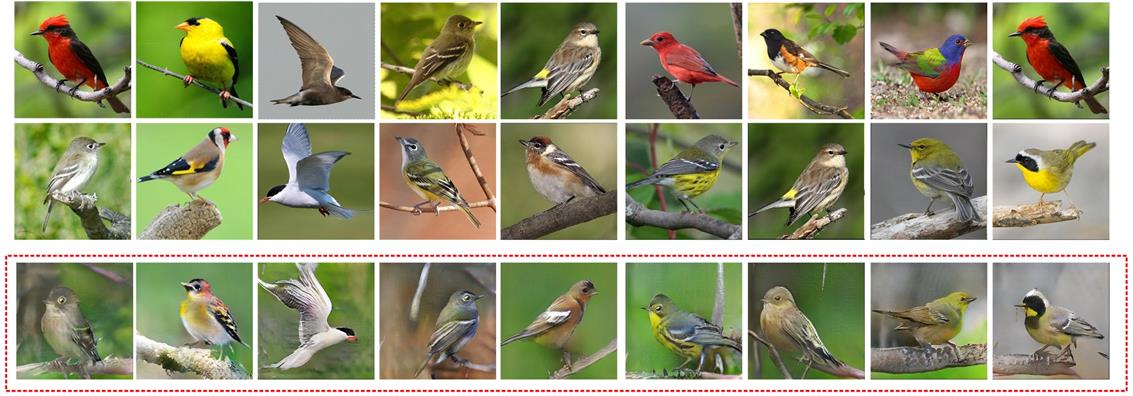}
	\caption{The first row is source images, and second row is target images. The translated images are placed in the third row, labeled by red dash box. }
	\label{config}
\end{figure*}

\vspace{-2mm}
More results generated by DA-GAN are shown in Figure \ref{text2img}. It can be seen that, given one description, the proposed DA-GAN is capable of generating diverse images according to the specific description. Comparing with existing text-to-image synthesis works, we train the DA-GAN by unpaired text-image data. Especially, because of our proposed \textbf{instance-level translation}, we can achieve high-resolution (256 $\times$ 256) images directly, which is more applicable than StackGAN (it needs two stages to achieve the same resolution). We also showed more results for Pose Morphing in Figure \ref{config}. Note that, the target should be bird breeds (image collections). Here we just random select one image to represent each bird breeds for reference.
\vspace{-3mm}


\subsubsection*{Human Face to Animation Face Translation}
\begin{figure}[t!]
	\centering
	\includegraphics[scale = 0.35]{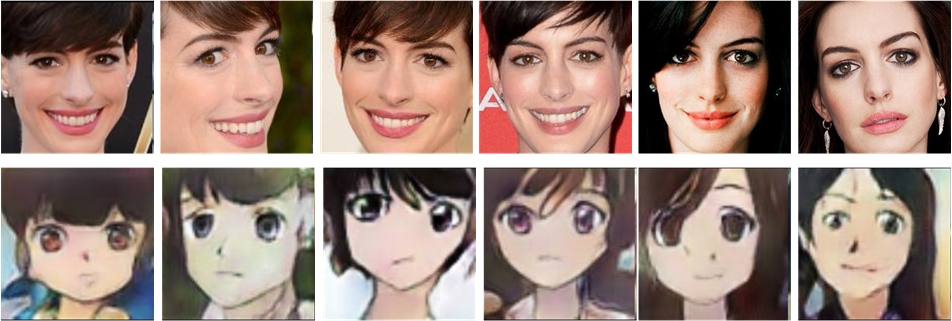}
	\caption{Fine-grained translation results.}
	\label{face_finegrain}
\end{figure}

\vspace{-2mm}
In this experiments, we randomly select 80 celebrities which consists of 12k images for source human face images. We also showed fine-grained translation results in Figure \ref{face_finegrain}. We can see that, with the same person, DA-GAN is capable of generating diverse images, while still remain the certain one's identity attributes, e.g. big round eyes, dark brown hairs, etc.

%
%

\subsubsection*{Translation on Paired Datasets}
\vspace{-2mm}
We also conduct experiments on paired datasets. The image quality of ours results is comparable to those produced by the fully supervised approaches while our method learns the mapping without paired supervision. For the task of Skeleton to cartoon figure translation, we retrieved about 20 cartoon figures which consists of 1200 images on websites, and adopt Pose Estimator by \cite{skeleton2016eccv} to generate skeletons for each image. The DA-GAN is trained by feeding into skeletons and generate cartoon images.

\begin{table} [t!]
	\begin{center}
		\begin{tabular}{|c|c|c|c|c|}
			\hline
			Datasets & Label $Y$ &  $\sharp$ & Instances & $d(\cdot)$ \\
			\hline\hline
			MNIST $\& $ SVHN & 10 & 1 & object & ResBlock \\
			\hline
			CUB-200-2011 & 200 & 4 & parts & VGG \\
			\hline
			FaceScrub & 80 & 4 & parts & Inception \\
			\hline
			Skeleton-cartoon & 20 & 4 & parts & VGG \\
			\hline
			CMP \cite{Tylecek2013} & None & 4 & parts & L2 \\
			\hline
			Colorization \cite{Jun:ICCV:2017:CycleGAN} & Binary & 1 & object & ResBlock \\
			\hline
			
		\end{tabular}
	\end{center}
	\vspace{-3mm}
	\caption{Implementation Details.}
	\label{detail}
	
\end{table}

\begin{figure*}
	\centering
	\includegraphics[scale=0.7]{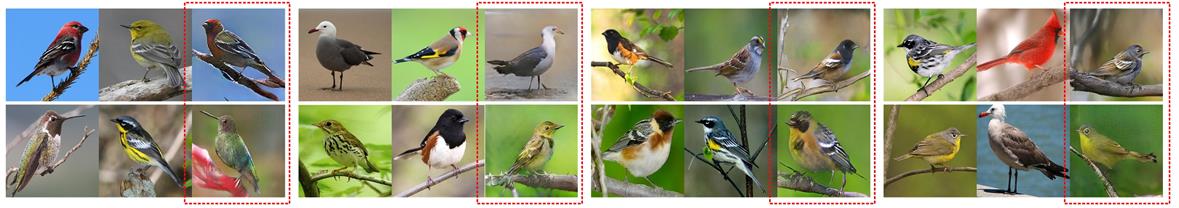}
	\caption{Results of pose morphing. In each group, the first column is the source image, the second row is target images. The red dashed box labeled the generated images, which possess the target objects pose while remain the source objects appearance.}
	\label{pose}
\end{figure*}

\begin{figure*}
	\centering
	\includegraphics[width=1.0 \textwidth]{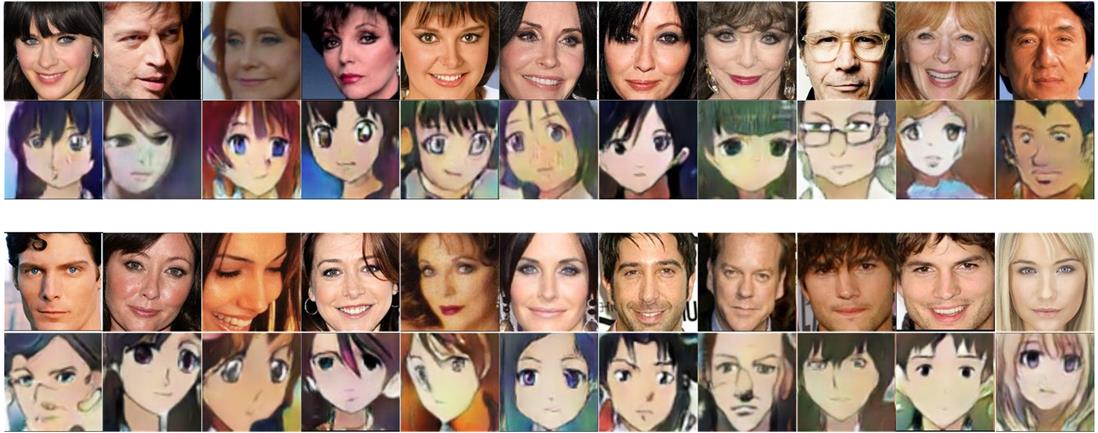}
	\caption{Results of human-to-animation faces translation. In each group, the first row is human faces, and the second row is translated animation faces.}
	\label{face}
\end{figure*}

\begin{figure*}
	\centering
	\includegraphics[width=1.0 \textwidth]{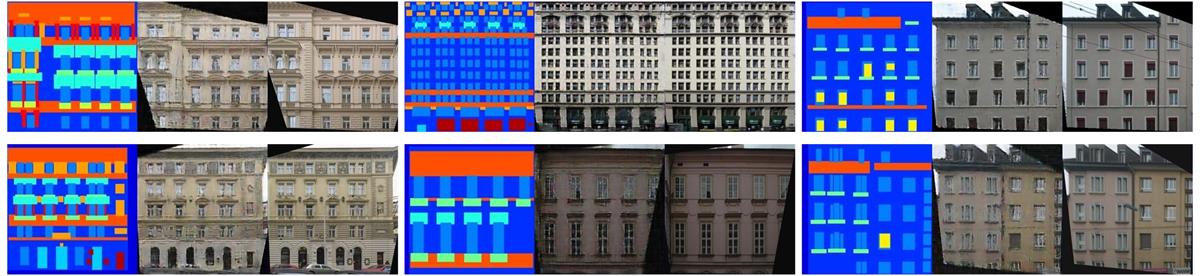}
	\caption{Results of architectural labels-to-photos translation. In each group from left to right are the input of labels, the translated architecture photos, and the ground truth.}
	\label{facades}
\end{figure*}

\begin{figure*}
	\centering
	\includegraphics[width=1.0 \textwidth]{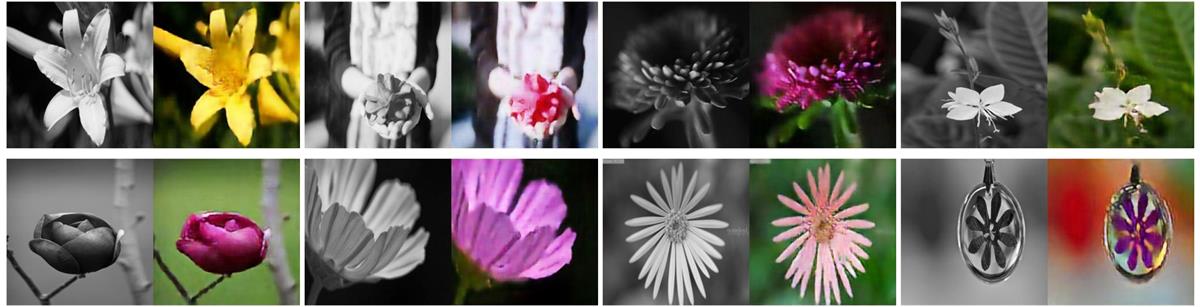}
	\caption{Results of image colorization. In each group, the input is gray images, and the results are translated color images.}
	\label{bw}
\end{figure*}

\end{document}